%% file: root.tex
\documentclass[sigconf,natbib=false]{aamas} 
\pdfoutput=1

\usepackage[boxed,ruled,vlined,linesnumbered]{algorithm2e}
\DontPrintSemicolon
\SetKwProg{Fn}{function}{}{}
\SetKwFunction{FnGetAction}{GetAction}
\SetKwFunction{FnDelegatePolicy}{DelegatePolicy}
\SetKwFunction{FnRestartArm}{RestartArm}
\SetKwFunction{FnStep}{Step}
\SetKwComment{Comment}{$\triangleright$\ }{}
\SetKwInput{KwInit}{Initialise}

\usepackage{etoolbox}
\makeatletter
\patchcmd{\@algocf@start}%
  {-1.5em}%
  {0pt}%
  {}{}%
\makeatother

\usepackage{wrapfig}
\usepackage{minibox}

\usepackage[inline]{enumitem}

\newcommand{\movetoappendex}[1]{{\color{green}#1}}

\renewcommand{\movetoappendex}{}

\usepackage{subcaption}

\usepackage{booktabs}
\usepackage{multirow}

\usepackage{amssymb}                                        %
\usepackage{mathtools}
\usepackage{braket}                                         %

\newcommand\numberthis{\addtocounter{equation}{1}\tag{\theequation}}

\newtheorem{definition}{Definition}

\usepackage{graphicx}                                       %
\usepackage{tikz}

\usepackage{bm}
\usepackage{dsfont}
\usepackage{xifthen}
\usepackage{stackengine} %
\usepackage{url}

\usepackage{cleveref}                                        %
\crefname{assumption}{assumption}{assumptions}
\crefname{problem}{problem}{problems}
\crefname{algorithm}{Alg.}{Algs.}
\Crefname{algorithm}{Algorithm}{Algorithms}
\crefname{figure}{Figure}{Figs.} %
\crefformat{equation}{(#2#1#3)}
\crefrangeformat{equation}{(#3#1#4) to~(#5#2#6)}
\crefmultiformat{equation}{(#2#1#3)}%
{ and~(#2#1#3)}{, (#2#1#3)}{ and~(#2#1#3)}

\crefformat{footnote}{#2\footnotemark[#1]#3}

\usepackage[boxed,ruled,vlined,linesnumbered]{algorithm2e}
\DontPrintSemicolon
\SetKwProg{Fn}{function}{}{}
\SetKwFunction{FnSampleFree}{SampleFree}
\SetKwFunction{FnRestartArm}{RestartArm}
\SetKwFunction{FnPickArm}{PickArm}
\SetKwFunction{FnRewire}{Rewire}
\SetKwFunction{FnNearest}{Nearest}
\SetKwFunction{FnRestartArm}{RestartArm}
\SetKwComment{Comment}{$\triangleright$\ }{}
\SetKwInput{KwInit}{Initialise}

\SetCommentSty{commentfont}

\input{_env_vars.tex}

\usepackage[
    backend=biber
    ,giveninits=true                  %
    ,url=false, isbn=false, doi=false %
    ,style=ieee
]{biblatex}

\usepackage{xpatch}
\xpatchbibmacro{date+extradate}{%
    \printtext[parens]%
}{%
    \setunit{\addperiod\space}%
    \printtext%
}{}{}

\DeclareFieldFormat[inproceedings]{title}{#1}   
\DeclareFieldFormat[incollection]{title}{#1}    
\DeclareFieldFormat[article]{title}{#1} 

\DefineBibliographyStrings{english}{%
    page             = {\ifbibliography{}{p\adddot}},
    pages            = {\ifbibliography{}{pp\adddot}},
} 

\DefineBibliographyExtras{english}{\def\finalandcomma{\addcomma}}
\DeclareDelimFormat{finalnamedelim}{\finalandcomma\addspace\bibstring{and}\space}

\DeclareNameAlias{sortname}{family-given}

\setcopyright{ifaamas}
\acmConference[AAMAS '21]{Proc.\@ of the 20th International Conference on Autonomous Agents and Multiagent Systems (AAMAS 2021)}{May 3--7, 2021}{London, UK}{U.~Endriss, A.~Now\'{e}, F.~Dignum, A.~Lomuscio (eds.)}
\copyrightyear{2021}
\acmYear{2021}
\acmDOI{}
\acmPrice{}
\acmISBN{}

\title{Robust Hierarchical Planning with Policy Delegation}

\acmSubmissionID{???}

\author{Tin Lai}
\affiliation{
  \department{School of Computer Science}
  \institution{The University of Sydney}}
\email{tin.lai@sydney.edu.au}

\author{Philippe Morere}
\affiliation{
  \department{School of Computer Science}
  \institution{The University of Sydney}}
\email{philippe.morere@sydney.edu.au}


\begin{abstract}
    We propose a novel framework and algorithm for hierarchical planning based on the principle of delegation. This framework, the \emph{Markov Intent Process}, features a collection of skills which are each specialised to perform a single task well. Skills are aware of their intended effects, and are able to analyse planning goals to delegate planning to the best suited skill. This principle dynamically creates a hierarchy of plans, in which each skill plans for sub-goals for which it is specialised.
    The proposed planning method features on-demand execution---skill policies are only evaluated when needed. Plans are only generated at the highest level, then expanded and optimised when the latest state information is available. The high-level plan retains the initial planning intent and previously computed skills, effectively reducing the computation needed to adapt to environment changes.
    We show this planning approach is experimentally very competitive to classic planning and reinforcement learning techniques on a variety of domains, both in terms of solution length and planning time.
\end{abstract}

\begin{document}

\pagestyle{fancy}
\fancyhead{}

\maketitle

\section{Introduction}
Decision making techniques enable automating many real-world tasks that would be too repetitive or even intractable to humans. Such methods require making good decisions in any situation, improved by reacting to the latest available information and revising previous intentions. This procedure is in practice extremely time sensitive, as the ability to react quickly is paramount in many real-world problems where the time available to make decisions is very limited.

Decision making typically involves generating plans by searching over the space of potential solutions. Classic planning methods search over plans by simulating various possible future scenarios. This quickly becomes prohibitively expensive in problems with larger state and/or action spaces, and greatly reduces their applicability to real-time problems.
Furthermore, plans generated this way often cannot be reused in similar situations, and unforeseen state changes often require expensive re-planning. This makes such planning techniques inefficient and slow.
More recent hierarchical planning and reinforcement learning methods overcome some of these issues by planning at several levels of abstraction.
Hierarchical planning decomposes the problem into smaller sub-problems, for which specialised skills are learned. These skills are highly reusable, easier to learn than general policies, and achieve better performance.
However, these hierarchical techniques are typically not aware of which task each skill is trained to solve. Knowledge of each skill's purpose is extremely important, as it allows seamlessly delegating planning to more specialised skills. This lack of awareness arises from the Markov Decision Process (MDP), the base framework for most of these methods, in which the effect of actions (or skills) is unknown. Because of this, existing hierarchical planning methods can be computationally inefficient, and learning specialised skills can prove challenging.

We present a hierarchical planning methodology for reasoning about the effects of skills and primitive actions, yielding several benefits. Planing using skill effect knowledge allows to directly select the best skill for any given task, reducing planning time and computation. This knowledge also enables planners to reason on whether executed skills or actions were successful by comparing the expected and observed effects; this allows learning action success conditions directly from interactions.
Furthermore, the presented method plans at the highest level only, and expands plans into more detailed plan \emph{on-demand}. Thus, latest state information can be taken into account, making plans reactive to noise and adversarial actions. Also, plans do \emph{not} need to be re-computed after unforeseen state changes occur, and planning computation is expended \emph{only} when a higher detail level is needed. This makes planning very innexpensive and fast.

Our contributions are the following.
We present a new sequential decision making framework, the \emph{Markov Intent Process} (MIP), incorporating action and skill effects at its core.
We formulate the notion of optimal plan in MIP, and propose to convert the sequential decision making problem into a collection of \emph{non-sequential} decision making problems, which are easier to solve.
We present a hierarchical intent-aware on-demand planning algorithm---called \emph{Delegate}---based on the MIP framework.
Finally, we experimentally show \emph{Delegate} is resilient to noise and outperforms other classic planning and reinforcement learning methods, both in terms of planning time and solution length, on a variety of domains.

\section{Related work}
\label{sec:related_work}
Planning by reasoning on the effects and conditions of actions has received much attention over the years. Classic planners like STRIPS~\autocite{fikes1971strips} are based on this principle, producing sequential plans using known action effects and conditions. Hierarchical Task Networks (HTN) improve on STRIPS by providing a hierarchical alternative to generate plans~\autocite{sacerdoti1975nonlinear,erol1996hierarchical,erol1994umcp,nau1999shop}. However, these planners necessitate skill hierarchy expert knowledge.
More work has aimed to extend HTN planners to learn the skill hierarchy automatically~\autocite{nejati2006learning}. However, skills hierarchy learning requires experts demonstrating solutions to the problem.

The field of reinforcement learning (RL) has also produced much work on sequential decision making~\autocite{sutton2018reinforcement}, mostly focusing on cases with unknown transition dynamics. RL defines the problem with a reward function, which agents aim to maximise, making most methods in this framework only applicable to a single goal.
When transition dynamics are learned from observed transitions, model-based RL methods~\autocite{sutton1991dyna} can generate optimal plans maximising rewards. These methods often leverage classic planners such as Monte-Carlo Tree Search~\autocite{coulom2006efficient} or model predictive control~\autocite{williams2017information} to simulate possible futures using the learned transition model. However, small errors in learned dynamics -- or even noise -- compound when planning over long horizons, making such approaches less robust.

RL has also benefited from the concept of hierarchical planning~\autocite{sutton1999between,dietterich2000hierarchical}.
The option framework~\autocite{sutton1999between} defines temporally extended actions called \emph{options} which can form hierarchical plans. Options are characterised by initiation and termination state sets, defining a region of the state space in which options can be used. This differs from MIP skills, which can all be executed in \emph{any} state, but aim at achieving very specific effects (e.g. setting the first state dimension to one).
While the nature and number of options is highly dependent on state space geometry~\autocite{csimcsek2005identifying}, desired state space manipulations (i.e. skill effects) are always known in advance, as defined by the action space; this is a stark advantage of skills over options.
Method to learn options by identifying transition data clusters was proposed in~\autocite{bakker2004hierarchical}, though it is limited to few levels of hierarchy.
Symbolic planning and reinforcement learning have also been combined in ~\autocite{grounds2008combining}, where the RL reward function is generated by a STRIPS. This approach combines reasoning advantages of STRIPS to the reactivity of RL.
Because RL is based on the MDP framework which defines objectives with a reward function, RL based methods typically can't handle multiple or changing goals.

Reasoning about actions and their effects is also the focus of relational RL~\autocite{van2005survey}, where state and action spaces are upgraded with objects and their relations. Following similar ideas, the Object-Oriented MDP framework~\autocite{diuk2008object} models state objects and their interactions. These ideas enable scaling RL to problems with very high dimensions, as long as they can be expressed in terms of a \emph{known} decomposition into objects.
Most similar to this work is that of~\autocite{morere2019learning}, which learns a hierarchy of skills using an expert-generated curriculum to guide learning. While the planning method is similar, it is restricted to specific sets of conditions and effects, thus not applicable to more complex problems.

This work combines the advantages of hierarchical planners to that of reasoning over action effects and conditions. Unlike previously mentioned work, the proposed method automatically delegates planning to the most specialised skill for the task, leading to faster and more efficient planning.

\section{Formulation}
We start by describing the Markov Intent Process---a novel framework for hierarchical decision making---that promotes delegating tasks whenever possible.

\subsection{Markov Intent Process}
\begin{definition}[Markov Intent Process]
A Markov Intent Process (MIP) is a tuple $\langle \mathcal{S}, \Skills, \Actions, E, T, \mathcal{C} \rangle$ composed of a set of states $\mathcal{S}$, a set of learnable (and initially empty) skills $\Skills$, a set of primitive actions $\Actions$ with known effects $e \in E$, a transition noise function $T$, and a set $\mathcal{C}$ of successful state spaces for each action.
a mip follows the Markov property, which states that the resulting state of a transition $s'$ only depends on the starting state $s$ and primitive action $\action \in \Actions$.
\end{definition}

\begin{definition}[Primitive action]
Each primitive action $\action \in \Actions$ is associated with a set of states $\mathcal{S}_{\action} \in \mathcal{C}$, in which executing $\action$ would be successful.
Successfully executing $\action$ in a state in $\mathcal{S}_{\action}$ would apply the action's effect $e$ to the current state.
This is given by the environment.
\end{definition}

\begin{definition}[High-level Skill]
A MIP skill $\skill_{e} \in \Skills$, identified by its effect $e$, is defined as a tuple $\skill_{e} := (e,\pi_{\skill_{e}})$ of
\begin{enumerate*}[label={(\roman*)}]
\item
    a \emph{known} intended effect $e$, describing the effect of the skill if $\skill_{e}$ is successfully executed (and no noise is observed); and
\item
    a learnable policy $\pi_{\skill_{e}}$ mapping a state in $\mathcal{S}$ to a plan 
    $\upsilon$.
\end{enumerate*}
Unlike actions, skills do not require a successful state space, as it can be automatically inferred from a skill's plan.
The set of skills $\Skills$ is \emph{not} given to MIP agents, which need to discover and learn them.
\end{definition}

\begin{definition}[plan]
A plan $\upsilon$, generated by a policy $\pi_{\skill_{e}}(s)$, is a (potentially empty) sequence of skills $\skill \in \Skills$ and/or actions $\action \in \Actions$ aiming to achieve effect $e$ from state $s$, possibly with some unknown side effects.
A plan implicitly defines a hierarchy and order of operations to achieve the skill's intended effect.
Hereafter, for notation brevity and clarity, we will refer to a skill's policy as $\pi_{e} \equiv \pi_{\skill_{e}}$.
\end{definition}

Each skill from the set of skills $\Skills$ has a known immutable effect. Skill policies may generate plans composed of a single primitive action $\action$, should it succeed in the current state $s$, i.e. $s \in \mathcal{S}_{\action}$. Otherwise, a skill's policy should return a plan composed of other skills, whose effects are necessary before the skill's effect could be achieved by a single primitive action from $\action$. As such, executing a skill may result in delegating parts of its plan to other skills.
That is, executing a plan might recursively queries other skill which in-turn returns another nested plan to be executed.

Given a state $s$ and a primitive action $\action$ with effect $e$, transitions to a new state $s'$ occur as:

\begin{equation}
    s' = \begin{cases}
    T(s \oplus e) \quad &\text{ if $s \in \mathcal{S}_{\action}$},\\
    T(s) \quad &\text{ otherwise}.
    \end{cases}
\end{equation}

Here, $\oplus$ denotes applying effect $e$ to state $s$, and the transition noise function $T$ potentially corrupts the resulting state $s'$ with noise.
The set of successful state spaces for all actions $\mathcal{C}$ and transition noise function $T$ are not necessarily known, thus MIP agents need to be reactive to the environmental changes.

\subsection{Action Success Conditions}
The MIP framework generalise the definition of the success conditions of an action $\action$ as the set of of states $\mathcal{S}_{\action}$ in which $\action$ succeeds. For simplicity, we only consider conditions that are expressed as the intersection of unit state space feature conditions.

Formally, let $\phi \colon \mathcal{S} \to \mathbb{R}$ denote one of $m$ functions mapping a state to a corresponding state feature.
Let $\mathcal{P}$ denote the space of propositions mapping a state feature $\phi(s)$ to a boolean value $\{true, false\}$.
We write $P^j_{\phi}(s)$ to denote a proposition $P^j \in \mathcal{P}$ operating on state feature $\phi$ in state $s$.

\begin{definition}[Action condition]
\label{def:conditions}
The condition of action $\action \in \Actions$ is defined as the minimal non-empty set of $d \le m$ feature propositions $\zeta_{\action} = \set{P^{j_1}_{\phi_{i_1}}(\cdot), \ldots, P^{j_d}_{\phi_{i_d}}(\cdot)}$, where $i_1, \ldots, i_d$ and $j_1, \ldots, j_d$ are all distinct,
such that
\begin{equation}
    \bigwedge_{P^j_{\phi_i}(\cdot) \in \zeta_{\action}} P^j_{\phi_i}(s) =
    \begin{cases}
        true &\text{ if $s \in \mathcal{S}_{\action}$},\\
        false &\text{ if $s \in \mathcal{S} \setminus \mathcal{S}_{\action}$}.
    \end{cases}
\end{equation}
\end{definition}
Intuitively, action condition $\zeta_{\action}$ from \cref{def:conditions} defines the state features that a sate $s$ must have for the action $\action$ to be successful.

\subsection{Well-formed Problems}
In order to guarantee MIPs are \emph{well-formed} and can be solved, we first need to define constraints on skills and action conditions, e.g. no circular dependency. Let us first define the notions of success prerequisite and well-formed action sets.
    
\begin{definition}[Success prerequisite]
We say that $\action_i$ is a \emph{success prerequisite} of action $\action_j$ 
in $s \not\in \mathcal{S}_{\action_j}$ if $\action_i$ is a necessary action in all possible action sequences that can transits $s$ to $s'\in \mathcal{S}_{\action_j}$.
\end{definition}
\begin{definition}[Well-formed action set]\label{def:well-formed-action-set}
    An action set $\Actions$ is said to be well-formed if it contains no ill-formed action; that is, no circular dependencies occur when unrolling conditions of $\action \in \Actions$.
    Formally, $\action$ is ill-formed if there exists a state $s \in \mathcal{S} \setminus \mathcal{S}_{\action}$ for which $\action$ itself is one of the success prerequisites.
\end{definition}

\movetoappendex{
Definition~\ref{def:well-formed-action-set} prevents any ill-formed actions form generating an illogical circular dependency, which would otherwise cause impossible situations where in order to reach states where $\action$ succeeds, action $\action$ must first be executed (i.e. $\action$ a prerequisite for itself).
A well-formed action set guarantees the conditions are logically sound; however, does not guarantee that the problem is solvable. To do so, we sart by defining the notion of sufficient action set.
\begin{definition}[Sufficient action set]\label{def:suff-action-set}
    Let $\mathcal{S}_0 \subseteq \mathcal{S}$ and $\mathcal{S}_g \subseteq \mathcal{S}$ be the sets of all possible start and goal states in our problem space respectively,
    $\Actions_\text{suff}$ be an action set that satisfies \cref{def:well-formed-action-set}, and $s_0 \in \mathcal{S}_0$ and  $s_g \in \mathcal{S}_g$.
    If there exists at least one action sequence $\upsilon = (\action_{e_i},\ldots,\action_{e_j})$ that can reach $s_g$ from $s_0$, where $\action_{e_i},\ldots,\action_{e_j} \in \Actions_\text{suff}$ and $e_i,\ldots,e_j \in E$, then $\Actions_\text{suff}$ is said to be a \emph{sufficient action set} for the problem space.
    Formally, denoting $\Upsilon_{\Actions_\text{suff}}$ as the space of all possible plans formed by using only elements in $\Actions_\text{suff}$,
    \begin{multline}
        \forall s_0 \in \mathcal{S}_0 \;\forall s_g \in \mathcal{S}_g \,
            \exists \upsilon = (\action_{e_i},\ldots,\action_{e_j}) \in \Upsilon_{\Actions_\text{suff}}, \\
        s_g = \left(\left(\left( s_0 \oplus e_i \right) \oplus \cdots \right)\oplus e_j \right).
    \end{multline}
\end{definition}

Notice that if we were to apply the transition noise function $T$ to the previous formulation, $\Actions_\text{suff}$ would no longer be sufficient, as a noisy transitions may result in states in which \cref{def:suff-action-set} does not hold;
i.e. there might exists $s' = T(s)$ where $s \in \mathcal{S}_0$ but $s' \not\in \mathcal{S}_0$, and there exists no plan $\upsilon \in \Upsilon_{\Actions_\text{suff}}$ such that $\upsilon$ can reach $s_g$ from $s$.
Therefore, $T$ could corrupt the current state into an unrecoverable status.
A stronger guarantee is that of complete action set.
\begin{definition}[Complete action set]\label{def:complete-action-set}
    An action set $\Actions_\text{comp}$ that satisfies \cref{def:well-formed-action-set} is said to be a \emph{complete action set} if it can complete all possible problems, even in the presence of noise $T$;
    i.e. if there exists at least one action sequence $\upsilon \in \Upsilon_{\Actions_\text{comp}}$ that can reach any $s_g \in \mathcal{S}$ from any $s_0 \in \mathcal{S}$.
\end{definition}
This is a special case of \cref{def:suff-action-set} where it reduces $\mathcal{S}_0 = \mathcal{S}_g = \mathcal{S}$, and it is immediate that $\Actions_\text{suff} \subseteq \Actions_\text{comp}$ as $\Actions_\text{comp}$ will always be \emph{sufficient} to complete any problem.
In practice, a \emph{sufficient action set} is often adequate if the noise function $T$ is relatively well behaved, e.g. in cases where $T$ can equally bring the state $s$ into unrecoverable state and bring it back to a favourable state space region.
In addition, we consider achieving goal state purely through noise as a trivial solution, which should not considered a possible solution.
Hereafter, $\Actions$ will be used to denote an action set that is at least as good as \emph{sufficient action set} (i.e. $\Actions_\text{suff} \subseteq \Actions$), and we consider
solutions to MIP problems should \emph{not} explicitly rely on noise.
}
Lastly, we restrict the arbitrary transition function $T$ to be a random and non-adversarial transition function.
That is because if an adversarial $T$ always negates any executed effects, there exists no plan that can make any progress.

\movetoappendex{
\subsection{Optimal Plans in a Markov Intent Process}

In the following, we will use $\length{\cdot}$ to denote the number of elements within a plan, 
and $\fulllength{\cdot}$ to denote the length of the \emph{full} plan (fully flattened plan where we recursively unroll all high-level skills), i.e.
\begin{equation}
    \fulllength{\upsilon_e} = \sum_{i=1}^{\length{\upsilon_e}} \fulllength{\upsilon_{e,i} },
\end{equation}
where $\upsilon_{e,i}$ is the $i^{th}$ element of $\upsilon_e$ corresponding to either an inner-plan or primitive action. If $\upsilon_{e,i}$ is a primitive action, $\fulllength{\upsilon_{e,i} }=1$.

Contrary to the MDP framework, no reward function is defined, thus solving a MIP cannot be defined as maximising rewards.
Instead, planning in a MIP requires a goal state $s_g \in \mathcal{S}_g$ to be specified. 
Solving a MIP optimally is equivalent to finding an optimal plan for all goal states in $\mathcal{S}_g$. 
The optimal plan for $s_g$ from a starting state $s_0$ is the shortest plan to $s_g$, given a limited planning horizon $H$. 
Formally, denoting $e$ the effect from $s_0$ to $s_g$, the optimal plan $\upsilon_e^{*H}$ is
\begin{equation}
    \label{eq:optimal_plan}
    \upsilon_e^{*H} = \arg\min_{\upsilon_e^{H}} \fulllength{\upsilon_e^{H}}.
\end{equation}
The planning horizon $H$ restricts the length of all plans within $\upsilon^H$ (and $\upsilon^H$ itself) to be no greater than $H$, i.e $\length{\upsilon_e^H } <= H$ and all plans within $\upsilon_e^H$ have horizon $H$.
For some short horizons and goals, a valid $\upsilon_e^H$ may not exist.

Because there is no fixed reward function (unlike in MDPs), MIPs agents must be able to generate plans for any given goal state. This is much harder than maximising a reward engineered for a single task, and makes MIPs naturally applicable to multi-task, transfer, and life-long learning tasks.
}

\section{Planning Through Delegation}
Planning optimally in a MIP requires solving Equation~\ref{eq:optimal_plan}, which involves an intractable search over all possible plans. We present instead a tractable method for approximating it, generating a successful on-demand plan for a given goal.

\subsection{Overview}
The proposed planning methodology is based on the principle of delegation; skills are defined such that they \emph{``Do one thing and do it well''}~\autocite{raymond2003_ArtUnix}. Each skill has a known effect $e$, and can generate a plan to achieve this effect from any state $s$. In the easiest situation, achieving $e$ in $s$ only involves returning a primitive action $\action$ whose effect is $e$. However, in most cases, $\action$ would fail if directly executed in $s$, and intermediate effects must first be applied to $s$ (through more planning) before $\action$ can be successfully executed. Once intermediate effects are identified, plans for each of them can be generated \emph{on demand} from other skills learned specifically for these effects. This lazy plan generation process enables automatically reacting to latest state changes (due to noise for example) and greatly reduces planning time.

\subsection{Formulation}
Formally, planning through delegation can be defined as follows.
Let each skill $\skill$ (with effect $e$) from the set of skills $\Skills$ be formulated as a MDP with no transition function nor discount factor $(\mathcal{S}, \Upsilon_e, R_e)$, denoted $\mathcal{M}_e$, following the contextual bandits framework~\autocite{auer2002nonstochastic}. The space of potential plans $\Upsilon_e$ to achieve $e$ contains at least all actions in $\Actions$ whose effect is $e$, and grows as more plans are evaluated by skill $\skill$. All plans in  $\Upsilon_e$ must have at most length $H$. The reward function $R_e$ is defined as follows:
\begin{equation}
\label{eq:bandit_reward}
    R_e(s, \upsilon) = \begin{cases}
                    \frac{1}{\fulllength{\upsilon}} &\text{ if executing $\upsilon$ in $s$ results in $e$},\\
                    0 &\text{ otherwise}.
                  \end{cases}
\end{equation}
Because skill effects are known, checking whether executing a plan from $s$ to $s'$ resulted in $e$ can be checked by inspecting whether $s'$ is consistent with $s \oplus e$. This can be done by the agent, without any additional knowledge (and without skill success condition knowledge).

Equation~\ref{eq:bandit_reward} promotes choosing plans with lower full plan length. Hence if the optimal plan $\upsilon^{*}$ for a given skill of effect $e$ is in $\Upsilon_e$, an agent maximising Equation~\ref{eq:bandit_reward} will generate an optimal plan for $e$, as defined in Equation~\ref{eq:optimal_plan}. If $\upsilon^{*} \not\in \Upsilon_e$, then more exploration is needed until $\upsilon^{*}$ is found. This problem can be solved as-is using contextual bandits solvers~\autocite{langford2007epoch,auer2002using,abe2003reinforcement}, in which a trade-off between exploration and exploitation is built-in. However, knowledge of skill success conditions, whether known in advance or learned through interactions, enables better reasoning over potential plans to consider and facilitates finding solutions.

\begin{algorithm}[tb]%
    \caption{Episode in MIP environment} \label{alg:mip-episode}
    \KwIn{$T$, $s_0$, $s_g$, maximum episode length $N$}
    Generate intent plan $\upsilon_0=(\skill_{e_0},\ldots,\skill_{e_i})$ from $s_0$ to $s_g$\;
    \For{$t\gets1$ \KwTo $N$}{
        $\action_e, \upsilon_t \gets \FnGetAction(\upsilon_{t-1}, s_{t-1})$ \;
        \If(\Comment*[f]{action can execute}){$s_{t-1} \in S_{\action_e}$ }{ 
            $s_t \gets s_{t-1} \oplus e$ \Comment*[r]{$e$ is effect of $\action$}
        }
        $s_t \gets T(s_t)$ \Comment*[r]{noisy environment only}
        \If{$s_t = s_g$}{
            \Return \emph{success}
        }
    }
    \Return \emph{failed}
\end{algorithm}

\subsection{Delegate}
We now describe how \emph{Delegate} leverages skill success conditions to efficiently find plans maximising Equation~\ref{eq:bandit_reward}.
While MIPs provide agents with knowledge of primitive action effects, there are two possible variants concerning success condition information: i) action success conditions are given to the agent, or ii) action success conditions are not given but agents can learn them through trial and error. In either approach, agents can construct plans by reasoning on the available primitive actions' effects and conditions. From here onward, we consider the case where conditions are known.

Planning in a MIP for a given goal state $s_g \in \mathcal{S}_g$ from a starting state $s_0 \in \mathcal{S}_0$ begins by identifying a sequence of effects ${e_0, \ldots, e_i}$ which results in $s_g$ if applied to $s_0$, i.e. $s_g = s_0 \oplus e_0 \oplus \ldots \oplus e_i$. Such sequence of effects must exist if the action space is complete. An initial plan $\upsilon_0$ for task $(s_0, s_g)$ is simply generated by retrieving the skills corresponding to these effects, $\upsilon_0 = (\skill_{e_0},\ldots,\skill_{e_i})$. This procedure happens whenever a planning starts with new goal state, as described in Algorithm~\ref{alg:mip-episode}.
This initial plan, called \emph{intent plan}, only contains high-level instructions; executing it will require further planning, which can be computed on demand.

\begin{algorithm}[tb]%
    \caption{Delegate algorithm} \label{alg:delegated-planning}
    \Fn{$\FnGetAction(\upsilon, s_t)$}{
        $x, \upsilon_\text{rest} \gets$ separate first element of $\upsilon$\;
        \uIf{$x \in \Actions$}{
            \Return{$x$, $\upsilon_\text{rest}$}
        }
        \Else(\Comment*[f]{$x \in \Skills$}){
            $\upsilon_\text{nested} \gets \FnDelegatePolicy(x, s_t)$ \;
            $\upsilon \gets \upsilon_\text{nested}:\upsilon_\text{rest}$ \Comment*[r]{concatenation}
            \Return{$\FnGetAction(\upsilon, s_t)$}
        }
    }
    
    \Fn{$\FnDelegatePolicy(\skill_e, s_t)$}{
        $\zeta_t \gets \left\{\; {P^j_{\phi_i}(\cdot) \in \zeta_{\action_e}} \given* \neg P^j_{\phi_i}(s_t) \;\right\}$ \Comment*[r]{unmet conditions}
    
        \uIf{$\zeta_t = \emptyset$}{ 
            \Return{$(\action_e)$} \Comment*[r]{terminal plan}
        }
        \Else{
            $\upsilon \gets \emptyset$\;
            \For(\Comment*[f]{for all unmet conditions}){${P^j_{\phi_i} \in \zeta_t}$}{
                $\skill_{e_j} \gets$ skill with effect that satisfies $P^j_{\phi_i}$ \;
                $\upsilon \gets \upsilon : \skill_{e_j}$ \Comment*[r]{concatenation}
            }
            \Return{$\upsilon$}
        }
    }
\end{algorithm}

\subsubsection{Skill delegation policy}
Once an intent plan $\upsilon_0$ is formed, generating actions to execute in specific states is achieved by querying the policies of skills within $\upsilon_0$. If the first element of $\upsilon_0$ is a primitive action, there's no need to call any skill policy, and the action can be executed directly. Else, if the first element of $\upsilon_0$ is a skill $\skill$, then its delegation policy $\pi_{\skill}$ is queried at the current state $s_t$. The resulting plan $\upsilon = \pi_{\skill}(s_t)$ is appended to the beginning of the intent plan $\upsilon_0$ (after $\skill$ was removed). The same process is applied to the new intent plan until its first element is a primitive action. This procedure is detailed in $\texttt{\FnGetAction}$, Algorithm~\ref{alg:delegated-planning}.

Generating a plan by querying a skill policy is achieved by analysing the success conditions $\zeta_{\action_e}$ of the skill's primitive action $\action_e$. The procedure is described in $\texttt{\FnDelegatePolicy}$, Algorithm~\ref{alg:delegated-planning}.
From Definition~\ref{def:conditions}, the set of conditions unmet in the current state can be computed $\zeta_t = \left\{\; {P^j_{\phi_i}(\cdot) \in \zeta_{\action_e}} \given* \neg P^j_{\phi_i}(s_t) \;\right\}$. If $\zeta_t$ is not empty, then we can construct a plan composed of skills whose effects would satisfy each of the individual elements $\zeta_t$. This plan is returned as the evaluation of the skill's policy in the current state.
In the event of $\zeta_t$ being an empty set, a terminal plan composed only of the skill's primitive action $\action_e$ is returned.
\begin{definition}[Terminal plan]
A plan $\upsilon$ is called a terminal plan if it contains no skills, i.e. $\forall x \in \upsilon, \nexists x \in \Skills$.
Empty plans where $\length{\upsilon} = 0$ are also considered terminal plans.
Terminal plans can be executed directly.
\end{definition}
Skill policies are queried recursively, until a termination criterion is met---when a terminal plan is generated.

\subsubsection{Planning termination}
Let us now analyse conditions under which planning terminates. As per Definition~\ref{def:well-formed-action-set}, a well-formed problem should contain no circular dependency.
However, a policy for skill $\skill$ could wrongly return a plan that contain skills that would generate plans containing $\skill$ itself. We call this plan \emph{ill-formed}, as it would result in circular dependencies.
\begin{definition}[Ill-formed plan]
A plan $\upsilon$ is said to be ill-formed if it contains skills that are ancestors of $\upsilon$. Formally, we define the direct parent of all skills within $\upsilon$ as skill $\skill_j$, such that plan $\upsilon$ was generated in some state $s_i$ as $\upsilon = \pi_{\skill_j}(s_i)$. Then $\upsilon$ is \emph{ill-formed} iff $\forall \skill \in \text{ancestors}(\upsilon), \exists \skill \in \upsilon$,
where ancestors are all parents of parents until the intent plan $\upsilon_0$, and the parent of a plan is equivalent to parent the skill that generated it.
\end{definition}
In practice, preventing ill-formed plan is achieved by imposing constraints on skill policies, such that no new plan includes any of its ancestors. Because the number of skills available to the planner is finite, the maximum plan depth is bounded. This guarantees a terminal plan will be generated eventually, hence ensuring planning termination.

\begin{figure*}[tb]
    \centering
    \includegraphics[width=\linewidth]{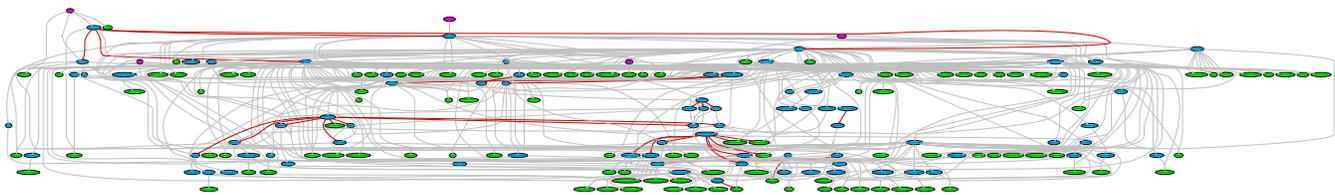}
    \caption{
        Full environment dynamics of the \emph{Factorio} domain%
        , representing dependency in between actions and states.
        \textcolor{magenta!90!black}{Magenta nodes} denote actions that have no conditions (i.e. the only entry point in the environment if the initial state is all zeroes),
        \textcolor{blue}{blue nodes} denote intermediate nodes and \textcolor{green!80!black}{green nodes} denote leaves in the graph (i.e. possible goals in the environment).
        \textcolor{gray!90!black}{Grey edges} denote actions with consuming effects (e.g. in~\cref{fig:factorio-environment-subgraph}, executing the action \skilltext{makeStoneFurance} will cancel state feature \statedimtext{hasStone} because it is used as material), and \textcolor{purple}{red edges} denote actions without such an effect (e.g. executing action \skilltext{makeStoneBrick} will not cancel \statedimtext{hasStoneFurance} because it is a tool).
        \label{fig:factorio-environment}
    }
\end{figure*}

\subsection{Analysis of a concrete example}
We now provide a concrete example of planning through delegation, illustrating the properties of \emph{Delegate}.

{

\begin{figure}
    \vspace{-10pt}
    \centering%
    \includegraphics[width=\linewidth]{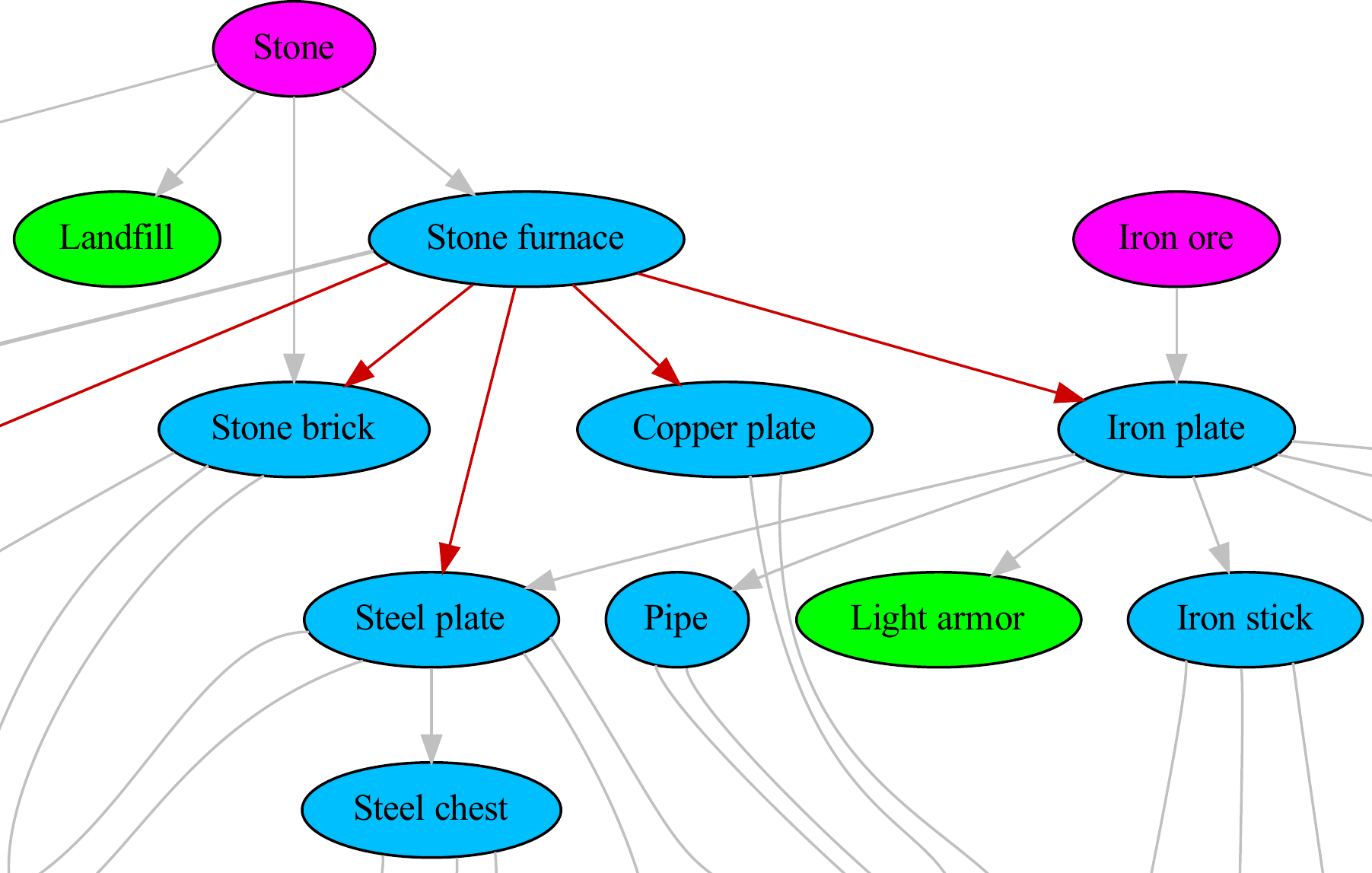}
    \caption{
         Cropped sub-graph of the \emph{Factorio} domain.
        Each node denotes a state and an action that has positive effect on that state;
        incoming edges of a node denote the conditions of that action.  
        For example, the action \skilltext{makeSteelPlate}
        is conditioned on \statedimtext{hasSteelPlate} being zero, and \statedimtext{hasStoneFurnace}, \statedimtext{hasIronPlate} being one;
        the action effect is cancelling state dimension \statedimtext{hasIronPlate} and satisfying \statedimtext{hasSteelPlate}.
        State dimension \statedimtext{hasSteelPlate} is in-turn conditioned by other actions such as \skilltext{makeSteelChest}.
        \label{fig:factorio-environment-subgraph}
    }
\end{figure}

In the following, we will use \statedimtext{hasState} to denote the presence of a state feature of interest, $\skill_{e:\statedimtext{hasState}}$ to denotes a skill with intended effect satisfying \statedimtext{hasState}, and $\action_\skilltext{doAction}$ to denote a primitive action with specified outcome.
We consider part of the Factorio environment,
represented as a graph in Figure~\ref{fig:factorio-environment-subgraph}. Skill $\skill_{e:\statedimtext{hasIronPlate}}$ must execute action $\action_\skilltext{makeIronPlate}$ to achieve its intended effect, however the action requires both state features \statedimtext{hasIronOre} and \statedimtext{hasStoneFurnace}.
Therefore, skill $\skill_{e:\statedimtext{hasIronPlate}}$ should form a plan that involves skills $\skill_{e:\statedimtext{hasIronOre}}$ and $\skill_{e:\statedimtext{hasStoneFurnace}}$, planning for effects activating missing state features.
Since $\action_\skilltext{getIronOre}$ has no conditions, the policy of skill $\skill_{e:\statedimtext{hasIronOre}}$ will return a terminal plan $\upsilon_{\skill_{e:\statedimtext{hasIronOre}}} = (\action_\skilltext{getIronOre})$, which can be executed directly.
However, $\action_\skilltext{makeStoneFurnace}$ is conditioned on \statedimtext{hasStone} and hence, skill $\skill_{e:\statedimtext{hasStoneFurnace}}$ should in turn generate a nested plan for $\skill_{e:\statedimtext{hasStone}}$.

}

A possible plan execution scenario for $\upsilon_{e:\statedimtext{hasSteelPlate}}$ is detailed below.
Note that we abbreviate ``\statedimtext{Stone}'' as ``\statedimtext{St}'', ``\statedimtext{Furnace}'' as ``\statedimtext{Fur}'', ``\statedimtext{Plate}'' as ``\statedimtext{Pl}'',  ``\statedimtext{make}'' as ``\statedimtext{mk}'' and ``\statedimtext{Iron}'' as ``\statedimtext{Ir}'' in the followings;
and we assume none of the conditions are met in the initial states $s_0$.

The element that is being unrolled at each line are are underlined, the $\becomes$ symbol denotes one possible policy, and the executed primitive action are in bold.
Notice that the scenario shown begins in state $s_0$, and every time an action is executed the environment transits to a new state $s_{i+1}$.
The new state is then used by the next skill's policy, until the environment finishes at $s_5$ (the state after executing the last action).
If the entire scenario is successful, the effect $e$:\statedimtext{hasIronPlate} is included in $s_5$.
Hence, each skill attempts to plan for the conditions that it requires, and delegates \emph{``how to achieve those conditions''} by querying the respective skill policies.
\begin{minipage}{\linewidth}
\begin{align*}
    &\underline{\pi_{e:\statedimtext{hasSteelPl}}(s_0)} \numberthis\label{eq:skill-hasSteelPl-policy}\\
    &\becomes (\underline{\skill_{e:\statedimtext{hasStFur}}}, \skill_{e:\statedimtext{hasIrPl}},\skill_{e:\statedimtext{hasSteelPl}}) \\
    &\becomes (\underline{\pi_{e:\statedimtext{hasStFur}}(s_0)}, \skill_{e:\statedimtext{hasIrPl}},\skill_{e:\statedimtext{hasSteelPl}}) \\
    &\becomes (\underline{(\skill_{e:\statedimtext{hasSt}}, \skill_{e:\statedimtext{hasStFur}})}, \skill_{e:\statedimtext{hasIrPl}},\skill_{e:\statedimtext{hasSteelPl}}) \\
    &\becomes ((\underline{\pi_{e:\statedimtext{hasSt}}(s_0)}, \skill_{e:\statedimtext{hasStFur}}), \skill_{e:\statedimtext{hasIrPl}},\skill_{e:\statedimtext{hasSteelPl}}) \\
    &\becomes (((\bm{\action_\textbf{\skilltext{getSt}}}), \underline{\pi_{e:\statedimtext{hasStFur}}(s_1)}), \skill_{e:\statedimtext{hasIrPl}},\skill_{e:\statedimtext{hasSteelPl}}) \\
    &\becomes (((\bm{\action_\textbf{\skilltext{getSt}}}), (\bm{\action_\textbf{\skilltext{mkStFur}}})), \underline{\pi_{e:\statedimtext{hasIrPl}}(s_2)},\skill_{e:\statedimtext{hasSteelPl}}) \numberthis\label{eq:skill-hasIrPl-policy} \\
    &\becomes (((\bm{\action_\textbf{\skilltext{getSt}}}), (\bm{\action_\textbf{\skilltext{makeStFur}}})), \underline{(\skill_{e:\statedimtext{hasIrOre}}, \skill_{e:\statedimtext{hasIrPl}})},\skill_{e:\statedimtext{hasSteelPl}}) \\
    &\becomes (((\bm{\action_\textbf{\skilltext{getSt}}}), (\bm{\action_\textbf{\skilltext{mkStFur}}})), (\underline{\pi_{e:\statedimtext{hasIrOre}}(s_2)}, \skill_{e:\statedimtext{hasIrPl}}),\skill_{e:\statedimtext{hasSteelPl}}) \\
    &\becomes (((\bm{\action_\textbf{\skilltext{getSt}}}), (\bm{\action_\textbf{\skilltext{mkStFur}}})), ((\bm{\action_\textbf{\skilltext{getIrOre}}}), \underline{\pi_{e:\statedimtext{hasIrPl}}(s_3)}),\skill_{e:\statedimtext{hasSteelPl}}) \\
    &\becomes (((\bm{\action_\textbf{\skilltext{getSt}}}), (\bm{\action_\textbf{\skilltext{mkStFur}}})), ((\bm{\action_\textbf{\skilltext{getIrOre}}}), (\bm{\action_\textbf{\skilltext{mkIrPl}}})),\underline{\pi_{e:\statedimtext{hasSteelPl}}(s_4)}) \\
    &\becomes (((\bm{\action_\textbf{\skilltext{getSt}}}), (\bm{\action_\textbf{\skilltext{mkStFur}}})), ((\bm{\action_\textbf{\skilltext{getIrOre}}}), (\bm{\action_\textbf{\skilltext{mkIrPl}}})),(\bm{\action_\textbf{\skilltext{mkSteelPl}}})) \\
\end{align*}
\end{minipage}

The element that is being unrolled at each line are are underlined, the $\becomes$ symbol denotes one possible policy, and the executed primitive action are in bold.
Notice that the scenario shown begins in state $s_0$, and every time an action is executed the environment transits to a new state $s_{i+1}$.
The new state is then used by the next skill's policy, until the environment finishes at $s_5$ (the state after executing the last action).
If the entire scenario is successful, the effect $e$:\statedimtext{hasIronPlate} is included in $s_5$.
Hence, each skill attempts to plan for the conditions that it requires, and delegates \emph{``how to achieve those conditions''} by querying the respective skill policies.

The role of a skill is thus to construct a plan that enables the associated primitive action to execute successfully.
That is, regardless of the current environment state, each skill is in charge of constructing a plan that can fulfil all necessary conditions before its primitive action can execute successfully.

\subsubsection{Reactivity to noise}
Although the given scenario did not include any noise, skills would automatically adapt to latest state changes if noise were present. Note how non-terminal plans generated by a skill always include the skill itself at the plan's last element. This rule plays a different role in the presence and absence of disrupting noise.
When no noise disrupts the plan, in Equation~\ref{eq:skill-hasSteelPl-policy}, skill $\skill_{e:\statedimtext{\textbf{hasSteelPl}}}$ generates a plan that includes itself as a skill in the last element (i.e. $\pi_{e:\statedimtext{hasSteelPl}}(s_0) = (\skill_{e:\statedimtext{hasStFur}}, \skill_{e:\statedimtext{hasIrPl}},\bm{\skill_{e:\statedimtext{\textbf{hasSteelPl}}}})$).
Typically, after all elements but last were executed (i.e. 
$(\bm{\ldots},\bm{\ldots},\allowbreak \skill_{e:\statedimtext{hasSteelPl}})$)
, the last skill generates a terminal plan composed of its primitive action, if all of its conditions are fulfilled at this stage (as is the case for all of the skills in the concrete example shown).

In the event of noise corrupting states, plans are dynamically adapted on the fly to correct, or even take advantage of the noise.
Suppose state dimension \statedimtext{hasStFur} was fulfilled, but subsequently cancelled by noise before the plan finished executing.
Then $s'_4 = T(s_4)$, and the executing $\pi_{e:\statedimtext{hasSteelPl}}$ in $s'_4$ results in plan $(\skill_{e:\statedimtext{hasStFur}},\allowbreak \skill_{e:\statedimtext{hasSteelPl}})$ because $\pi_{e:\statedimtext{hasSteelPl}}$ automatically corrects the plan to fulfil the cancelled state feature \statedimtext{hasStFur}.
This property is also very valuable when noise enables one of the state features that will be required to execute the plan. In this case, parts of the plan would be skipped altogether, hence only executing actions that would succeed.

\subsubsection{Benefit of the decoupled approach}
Planning through delegation helps decoupling knowledge of different tasks, by formalising each skill as a separate MDP specialised for a single task. 
Aside from making it easier to plan for a single objective, it helps reducing redundant actions by mitigating possible action overlaps when planning for multiple effects.
An example is shown in Equation~\ref{eq:skill-hasIrPl-policy}, where the policy $\pi_{e:\statedimtext{hasIrPl}}$ constructs a plan that does not include skill $\skill_{e:\statedimtext{hasStFur}}$ even though \statedimtext{hasStFur} is one of the required conditions for $\skill_{e:\statedimtext{hasIrPl}}$ (see Figure~\ref{fig:factorio-environment-subgraph}).
This is explained by $\pi_{e:\statedimtext{hasSteelPl}}(s_0)$ generating a plan that includes $\skill_{e:\statedimtext{hasStFur}}$ in Equation~\ref{eq:skill-hasSteelPl-policy}. Hence when policy $\pi_{e:\statedimtext{hasIrPl}}(s_2)$ is queried in Equation~\ref{eq:skill-hasIrPl-policy}, \statedimtext{hasStFur} is already satisfied in $s_2$, and there is no need to execute $\skill_{e:\statedimtext{hasStFur}}$ again.

\section{Experiments}
The proposed planning method is now experimentally evaluated on a variety of domains of increasing complexity. We provide code for the algorithm, domains and baselines.%
We first describe the environments and experimental setup, then analyse the proposed method's sensitivity to noise, and finally compare it to several baselines.

\begin{table}[bt]
    \tabcolsep=0.11cm
    \resizebox{\linewidth}{!}{%
    \input{table-env-stats}
    }
    \caption{Properties of environments tested in experiments
    \label{tab:env_description}}
\end{table}

\subsection{Environments}
Environments are modelled as directed acyclic dependency graphs, with nodes being state features (i.e. the value $0$ or $1$ of each node is a feature) and edges representing actions and their success conditions. We run comparisons on five different environments with different properties, as shown in Table~\ref{tab:env_description}. Environments \emph{Mining}, \emph{Crafting} and \emph{Random} are as presented in~\autocite{morere2019learning}, where all effects turn a single state feature from $0$ to $1$. \emph{MiningV2} is a variant of \emph{Mining} with more complex effects -- consuming effects -- which turn some state features from $1$ back to $0$. For example, the effect of crafting an iron pickaxe consumes a stick and iron, i.e. \statedimtext{hasIronPickaxe} changes from $0$ to $1$ and \statedimtext{hasIron} and \statedimtext{hasStick} change from $1$ to $0$. 
Lastly, we present the more complex environment \emph{Factorio}, directly generated from the crafting and construction dependency of a video game\footnote{\path{https://wiki.factorio.com}}. 
This environment features consuming effects and even more complex effects toggling several state features simultaneously.

\begin{table*}[bt]
\resizebox{\linewidth}{!}{%
\input{table-expr-result}
}
\caption{Experimental results from various methods and environments ($\mu\pm\sigma$ over 10 runs)
    \label{table:expr-result}}
\end{table*}

\movetoappendex{

\subsection{Experimental setup}

The proposed method is experimentally compared to the following baselines.
\emph{Monte-Carlo Tree Search} (MCTS)~\autocite{coulom2006efficient} is a planning method that builds a (non-exhaustive) tree of possible futures, executing the action with the best long-term score (defined as a reward function of the goal state). MCTS was chosen as a baseline for its successes in robotics and RL.
The amount of search is defined by a budget; experiments analyse values of $100$, $1000$, and $5000$. The search is performed at every time-step after an action is executed. MCTS exploration constant is set to $\frac{1}{\sqrt{2}}$.
\emph{Rapidly-exploring Random Trees} (RRT)~\autocite{lavalle1998rapidly} is another sample-based planner which grows an expanding tree from starting state to free space. RRT can enhance its search by biasing it towards the goal state with low probability ($0.05$). In experiments, we use maximum allowed tree sizes of $1000$ and $10000$. 
The performance of RRT is bounded by the size of its search tree, where our experiments tested $1000$ and $10000$ as the maximum node counts.
Note that a search tree is built at every time-step to re-plan for any state changes and noises.
\emph{Q-Learning}~\autocite{watkins1992q} is a RL method that requires training to enhance its performance. It is guaranteed to converge to the optimal policy, given enough time~\autocite{jaakkola1994convergence}. It can only handle planning for a single goal, defined in a reward function. Unlike other baselines, it does not have access to transition dynamics, although these are implicitly learned during training. We used a discount factor $\gamma=0.99$, a learning rate of $0.1$ and an $\epsilon$-greedy policy with random action probability $0.1$. A reward of $0$ is given for reaching the goal, and $-1$ rewards are given otherwise. Q-values are tabular and initialised to $0$, ensuring exploration by making the agent optimistic. The method is trained for $20{,}000$ training episodes (typically less than 1 hour).
\emph{Hierarchical}~\autocite{morere2019learning} is a hierarchical planner that constructs plans of user-defined skills, similarly to the proposed method. Planning is achieved by planning backwards from goal to state. The method, however, can only handle non-consuming effects.
All experiments are averaged over $10$ runs on single core $2.2$GHz. We report planning success rates, i.e. number of times the agent manages to reach the given goal. Plan length and planning times are reported for successful runs \emph{only}.

All planners are tested against environments that are noise-free and with noisy-transition $T(\cdot)$.
A noise probability of $0.05$ is chosen, which denotes the probability of one dimension inverting its value every time the agent executes an action (regardless of whether the action is valid or not).

}

\begin{figure}[tb]
    \centering
    \includegraphics[width=\linewidth]{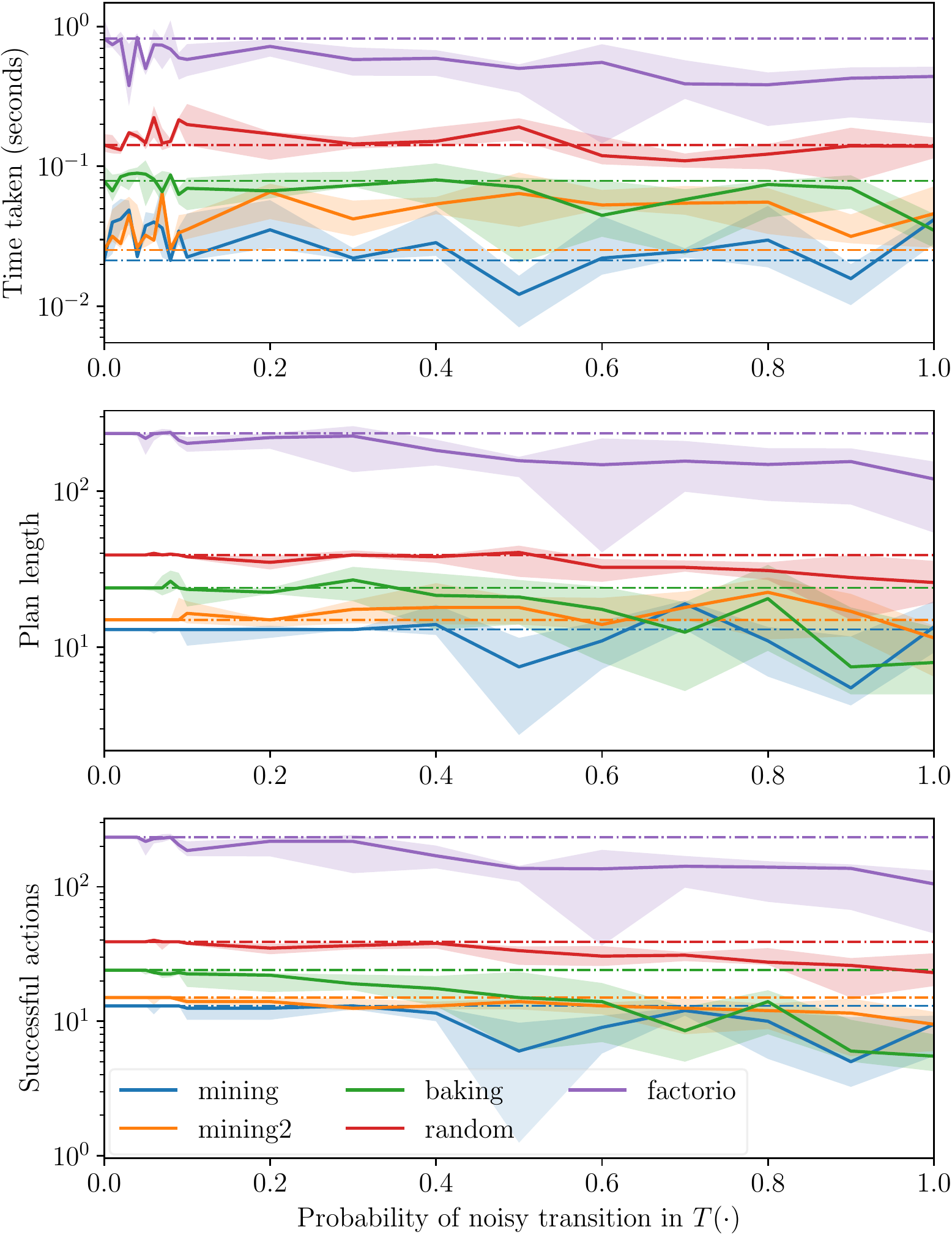}
    \caption{
        \emph{Delegate} performance (median, $25\%$ and $75\%$ quartiles) on all domains for different levels of noise.
        \label{fig:noise-sensitivity}
    }
\end{figure}

\subsection{Sensitivity to noise}
We begin by analysing the effect of transition noise on the proposed planner. All environments are evaluated with transition noise changing a single state feature from $0$ to $1$ or $1$ to $0$ with different probabilities ranging from $0$ to $1$. Results displayed in Figure~\ref{fig:noise-sensitivity} show the method is almost unaffected by noise.
In more complex environments, the noise is less likely to tamper state features affected by the current section of the plan than it is to disturb any other state feature. As such, noise tends to be more beneficial in complex environments, as shown by decreased plan length. The opposite happens in easier environments. The number of successful actions also decreases, as noise effects are sometimes aligned with the plan. Planning times only mildly decrease with noise. 

\subsection{Results}
We now compare the proposed method to baselines on all environments, with and without noise. Results are provided in Table~\ref{table:expr-result}.
Note that RRT is not applicable in the Factorio environment because it requires a one-to-one mapping from action effect to state dimension to expand its tree.
Because Q-Learning builds a mapping between state-action pairs and their Q values, it ran out of memory (64 GB) after experiencing too many different pairs (e.g. in the Factorio environment).
The Hierarchical planner is not applicable to MiningV2 and Factorio because it cannot reason on environment with consuming effects.

Overall the proposed method \emph{Delegate} consistently outperforms other baselines on all environments, whereas other methods are very sensitive to increasing environment complexity and noise.
In particular, methods that do not reason on action conditions don't succeed in complex environments (e.g. Random and Factorio).
Q-learning achieves near-optimal plan length on Mining, as the problem dimension is small enough. However, Q-learning is quickly overwhelmed by increasing problem sizes, and it does not converge in the allocated training time or even runs out of memory.
Similar results are observed in MCTS and RRT, however, the solutions obtained are far from optimal.
The two methods also requires explicit re-planning at each time-step to account for state changes and noise.
As a result, they require lengthy runtime to complete a single episode.
\emph{Hierarchical} is able to achieves near-optimal result in most applicable environments, and its execution speed is slightly lower than \emph{Delegate}.
However, in contrast to \emph{Delegate} it is susceptible to noise and complex environments as the methods plans out a horizon of action sequence, which could be invalidated by environment noise.

\section{Conclusion}
We proposed a framework and method for planning through delegation, by dividing the sequential decision making problem into multiple non-sequential problems with one specific goal each.
Each skill only plans for \emph{what} it needs and \emph{when} it needs it, delegating how to achieve necessary sub-goals to the most specialised skill. The presented planner generates plans on-demand, making it resilient to noisy transitions and able to adapt to latest state changes. We experimentally show it outperforms other classic planner and RL methods on a variety of environments.

As future work, the method could be extended to unknown action conditions, which would be learned from observing action successes and failures.
By subdividing the planning task into several simpler problems, learning spares conditions would be relatively easier.
Lastly, a model-free approach based on classic contextual bandits could also be considered.

\printbibliography

\end{document}

%% file: _env_vars.tex
\definecolor{amethyst}{rgb}{0.6, 0.4, 0.8}
\definecolor{alizarin}{rgb}{0.82, 0.1, 0.26}
\definecolor{ashgrey}{rgb}{0.43, 0.5, 0.5}
\definecolor{yellow}{rgb}{1.0, 0.75, 0.0} %

\makeatletter
\newcommand{\overrightsmallarrow}{\mathpalette{\overarrowsmall@\rightarrowfill@}}
\newcommand{\overarrowsmall@}[3]{%
  \vbox{%
    \ialign{%
      ##\crcr
      #1{\smaller@style{#2}}\crcr
      \noalign{\nointerlineskip}%
      $\m@th\hfil#2#3\hfil$\crcr
    }%
  }%
}
\def\smaller@style#1{%
  \ifx#1\displaystyle\scriptstyle\else
    \ifx#1\textstyle\scriptstyle\else
      \scriptscriptstyle
    \fi
  \fi
}
\makeatother

\mathchardef\ordinarycolon\mathcode`\:
\mathcode`\:=\string"8000
\begingroup \catcode`\:=\active
  \gdef:{\mathrel{\mathop\ordinarycolon}}
\endgroup

\usepackage{mathtools}

\makeatletter
\newcommand{\@givenstar}{\;\middle|\;}
\newcommand{\@givennostar}[1][]{\;#1|\;}
\newcommand{\given}{\@ifstar\@givenstar\@givennostar}
\makeatother

\newcommand{\action}{\hat{a}}
\newcommand{\skill}{\alpha}
\newcommand{\Actions}{\hat{\mathcal{A}}}
\newcommand{\Skills}{\mathcal{A}}

\newcommand{\length}[1]{{\mid #1 \mid}}
\newcommand{\fulllength}[1]{{\parallel #1 \parallel}}

\newcommand{\skilltext}[1]{\texttt{#1}}
\newcommand{\statedimtext}[1]{\textsl{#1}}

\DeclareFontFamily{U} {MnSymbolA}{}
\DeclareFontShape{U}{MnSymbolA}{m}{n}{
  <-6> MnSymbolA5
  <6-7> MnSymbolA6
  <7-8> MnSymbolA7
  <8-9> MnSymbolA8
  <9-10> MnSymbolA9
  <10-12> MnSymbolA10
  <12-> MnSymbolA12}{}
\DeclareFontShape{U}{MnSymbolA}{b}{n}{
  <-6> MnSymbolA-Bold5
  <6-7> MnSymbolA-Bold6
  <7-8> MnSymbolA-Bold7
  <8-9> MnSymbolA-Bold8
  <9-10> MnSymbolA-Bold9
  <10-12> MnSymbolA-Bold10
  <12-> MnSymbolA-Bold12}{}
\DeclareSymbolFont{MnSyA} {U} {MnSymbolA}{m}{n}
\DeclareMathSymbol{\rhookrightarrow}{\mathrel}{MnSyA}{48}
\newcommand{\becomes}{%
    \scalebox{.8}[1]{$\rhookrightarrow$}
}

%% file: table-env-stats.tex
\begin{tabular}{@{}lcccccc@{}}
\toprule
\multicolumn{1}{c}{Env.} & \begin{tabular}[c]{@{}c@{}}Consuming\\ effect\end{tabular} & \begin{tabular}[c]{@{}c@{}}Ep.\\ length\end{tabular} & \begin{tabular}[c]{@{}c@{}}Num.\\ actions\end{tabular} & \begin{tabular}[c]{@{}c@{}}Num.\\ nodes\end{tabular} & \begin{tabular}[c]{@{}c@{}}Avg. edges\\ per node\end{tabular} & \begin{tabular}[c]{@{}c@{}}State space\\ size\end{tabular} \\ \midrule
Mining                   & No                                                         & 40                                                   & 22  & 22                                                   & $2.27\pm1.14$                                                 & $4.19\times10^{6}$                                         \\
MiningV2                 & Yes                                                        & 80                                                   & 22    & 22                                                   & $3.18\pm1.97$                                                 & $4.19\times10^{6}$                                         \\
Baking                   & No                                                         & 60                                                   & 30    & 30                                                   & $2.00\pm0.37$                                                 & $1.07\times10^{9}$                                         \\
Random                   & No                                                         & 100                                                  & 100    & 100                                                  & $1.32\pm0.71$                                                 & $1.27\times10^{30}$                                        \\
Factorio                 & Yes                                                        & 300                                                  & 193     & 194                                                & $6.10\pm15.45$                                                & $2.51\times10^{58}$                                        \\ \bottomrule
\end{tabular}

%% file: table-expr-result.tex
\begin{tabular}{@{}p{1.2cm}p{.8cm}@{}ccccccccc@{}}
\toprule
                          & \begin{tabular}[c]{@{}c@{}}$T(\cdot)$\\ prob.\end{tabular} &                                              & Q-Learning         & \begin{tabular}[c]{@{}c@{}}MCTS\\ (100)\end{tabular} & \begin{tabular}[c]{@{}c@{}}MCTS\\ (1000)\end{tabular} & \begin{tabular}[c]{@{}c@{}}MCTS\\ (5000)\end{tabular} & \begin{tabular}[c]{@{}c@{}}RRT\\ (1000)\end{tabular} & \begin{tabular}[c]{@{}c@{}}RRT\\ (10000)\end{tabular} & Hierarchical    & PolicyDelegate    \\ \midrule
\multirow{6}{*}{Mining}   & \multirow{3}{*}{0.00}                                      & \multicolumn{1}{p{1.1cm}|}{Success}                 & $80.0 \%$          & $30\%$                                               & $70 \%$                                               & $80.0\%$                                              & $0.0 \%$                                             & $40.0 \%$                                             & $\bm{100\%}$         & $\bm{100\%}$           \\
                          &                                                            & \multicolumn{1}{p{1.1cm}|}{$\fulllength{\upsilon}$} & $13.25 \pm 0.5$    & $38.67 \pm 1.155$                                    & $35.14 \pm 4.947$                                     & $34.25\pm4.132$                                       & --                                                   & $32.25 \pm 4.425$                                     & $\bm{13.0}\pm0.0$    & $\bm{13.0} \pm 0.0$    \\
                          &                                                            & \multicolumn{1}{p{1.1cm}|}{Time}                    & $0.024 \pm 0.039$  & $21.94 \pm 2.659$                                    & $119.9 \pm 19.52$                                     & $184.4\pm46.39$                                       & $11.01 \pm 0.207$                                    & $65.71 \pm 6.534$                                     & $\bm{0.007}\pm0.001$ & $0.028\pm0.008$   \\ \cmidrule(l){2-11} 
                          & \multirow{3}{*}{0.05}                                      & \multicolumn{1}{p{1.1cm}|}{Success}                 & $50.0 \%$          & $10 \%$                                              & $10 \%$                                               & $20.0\%$                                              & $60.0 \%$                                            & $40.0 \%$                                             & $\bm{100\%}$         & $\bm{100\%}$          \\
                          &                                                            & \multicolumn{1}{p{1.1cm}|}{$\fulllength{\upsilon}$} & $13.0 \pm 0.0$     & $33.0 \pm 0.0$                                       & $38.0 \pm 0.0$                                        & $37.0\pm1.414$                                        & $31.17 \pm 9.304$                                    & $31.75 \pm 1.893$                                     & $13.1\pm0.3162$ & $\bm{12.65} \pm 1.565$ \\
                          &                                                            & \multicolumn{1}{p{1.1cm}|}{Time}                    & $0.023 \pm 0.024$  & $21.94 \pm 2.659$                                    & $119.9 \pm 19.52$                                     & $80.73\pm10.28$                                       & $9.862 \pm 2.325$                                    & $72.44 \pm 10.77$                                     & $\bm{0.009}\pm0.002$ & $0.031 \pm 0.016$ \\ \midrule
\multirow{6}{*}{MiningV2} & \multirow{3}{*}{0.00}                                      & \multicolumn{1}{p{1.1cm}|}{Success}                 & $0.0 \%$           & $90 \%$                                              & $\bm{100\%}$                                              & $\bm{100\%}\%$                                             & $0.0 \%$                                             & $0.0 \%$                                              & N/A             & $\bm{100\%}$          \\
                          &                                                            & \multicolumn{1}{p{1.1cm}|}{$\fulllength{\upsilon}$} & --                 & $47.33 \pm 11.03$                                    & $34.2 \pm 5.051$                                      & $37.1\pm8.212$                                        & --                                                   & --                                                    & --              & $\bm{15.0} \pm 0.0$    \\
                          &                                                            & \multicolumn{1}{p{1.1cm}|}{Time}                    & $0.131 \pm 0.051$  & $55.79 \pm 15.36$                                    & $241.1 \pm 86.49$                                     & $362.4\pm97.02$                                       & $21.25 \pm 0.885$                                    & $136.8 \pm 0.486$                                     & --              & $\bm{0.028}\pm0.007$   \\ \cmidrule(l){2-11} 
                          & \multirow{3}{*}{0.05}                                      & \multicolumn{1}{p{1.1cm}|}{Success}                 & $40.0 \%$          & $30 \%$                                              & $70 \%$                                               & $70.0\%$                                              & $20.0 \%$                                            & $50.0 \%$                                             & N/A             & $\bm{100\%}$          \\
                          &                                                            & \multicolumn{1}{p{1.1cm}|}{$\fulllength{\upsilon}$} & $107.2 \pm 68.88 $ & $43.33 \pm 4.509$                                    & $56.71 \pm 22.34$                                     & $65.43\pm12.92$                                       & $73.5 \pm 9.192 $                                    & $25.6 \pm 23.37 $                                     & --              & $\bm{14.4} \pm 2.683$  \\
                          &                                                            & \multicolumn{1}{p{1.1cm}|}{Time}                    & $0.080 \pm 0.031$  & $55.79 \pm 15.36$                                    & $241.1 \pm 86.49$                                     & $215.7\pm60.7$                                        & $21.63 \pm 1.193$                                    & $91.95 \pm 57.03$                                     & --              & $\bm{0.041} \pm 0.021$ \\ \midrule
\multirow{6}{*}{Baking}   & \multirow{3}{*}{0.00}                                      & \multicolumn{1}{p{1.1cm}|}{Success}                 & $0.0 \%$           & $0 \%$                                               & $70 \%$                                               & $80.0\%$                                              & $\bm{100\%}$                                           & $\bm{100\%}$                                            & $\bm{100\%}$       & $\bm{100\%}$          \\
                          &                                                            & \multicolumn{1}{p{1.1cm}|}{$\fulllength{\upsilon}$} & --                 & --                                                   & $49.86 \pm 3.132$                                     & $47.62\pm5.012$                                       & $37.2 \pm 3.706$                                     & $28.2 \pm 0.919$                                      & $\bm{24.0}\pm0.0 $   & $\bm{24.0} \pm 0.0$    \\
                          &                                                            & \multicolumn{1}{p{1.1cm}|}{Time}                    & $0.101 \pm 0.012$  & $32.27 \pm 14.08$                                    & $178.3 \pm 56.21$                                     & $943.3\pm219.3$                                       & $10.93 \pm 1.042$                                    & $53.93 \pm 2.449$                                     & $\bm{0.012}\pm0.002$ & $0.047\pm0.009$   \\ \cmidrule(l){2-11} 
                          & \multirow{3}{*}{0.05}                                      & \multicolumn{1}{p{1.1cm}|}{Success}                 & $20.0 \%$          & $40 \%$                                              & $20 \%$                                               & $30.0\%$                                              & $\bm{100\%}$                                           & $\bm{100\%}$                                            & $\bm{100\%}$       & $\bm{100\%}$          \\
                          &                                                            & \multicolumn{1}{p{1.1cm}|}{$\fulllength{\upsilon}$} & $43.0 \pm 46.67$   & $27.5 \pm 21.38$                                     & $\bm{22.0} \pm 15.56$                                      & $50.0\pm13.08$                                        & $33.7 \pm 13.05$                                     & $29.0 \pm 1.886$                                      & $22.6\pm8.003$  & $22.6 \pm 5.016$  \\
                          &                                                            & \multicolumn{1}{p{1.1cm}|}{Time}                    & $0.075\pm 0.036$   & $32.27 \pm 14.08$                                    & $178.3 \pm 56.21$                                     & $138.6\pm20.78$                                       & $9.986 \pm 3.943$                                    & $57.45 \pm 4.622$                                     & $\bm{0.017}\pm0.006$ & $0.052 \pm 0.019$ \\ \midrule
\multirow{6}{*}{Random}   & \multirow{3}{*}{0.00}                                      & \multicolumn{1}{p{1.1cm}|}{Success}                 & $0.0\%$            & $0 \%$                                               & $0 \%$                                                & $0.0\%$                                               & $0.0 \%$                                             & $0.0 \%$                                              & $70.0\%$        & $\bm{100\%}$          \\
                          &                                                            & \multicolumn{1}{p{1.1cm}|}{$\fulllength{\upsilon}$} & --                 & --                                                   & $--$                                                  & --                                                    & --                                                   & --                                                    & $66.57\pm23.02$ & $\bm{39.0} \pm 0.0$    \\
                          &                                                            & \multicolumn{1}{p{1.1cm}|}{Time}                    & $1.028\pm0.003$    & $206.5 \pm 36.05$                                    & $1745 \pm 478.9$                                      & $6564\pm23.47$                                        & $41.82 \pm 1.724$                                    & $275.7 \pm 3.028$                                     & $\bm{0.084}\pm0.037$ & $0.104\pm0.044$   \\ \cmidrule(l){2-11} 
                          & \multirow{3}{*}{0.05}                                      & \multicolumn{1}{p{1.1cm}|}{Success}                 & $0.0\%$            & $10 \%$                                              & $10 \%$                                               & $0.0\%$                                               & $0.0 \%$                                             & $10.0 \%$                                             & $60.0\%$        & $\bm{100\%}$          \\
                          &                                                            & \multicolumn{1}{p{1.1cm}|}{$\fulllength{\upsilon}$} & --                 & $47.0 \pm 0.0$                                       & $\bm{25.0} \pm 0.0$                                        & --                                                    & --                                                   & $9.0 \pm 0.0$                                         & $63.33\pm15.6$  & $35.9 \pm 9.457$  \\
                          &                                                            & \multicolumn{1}{p{1.1cm}|}{Time}                    & $0.902\pm0.175$    & $206.5 \pm 36.05$                                    & $1745 \pm 478.9$                                      & $3383\pm269.2$                                        & $45.56 \pm 2.103$                                    & $213.7 \pm 67.2$                                      & $0.091\pm0.031$ & $\bm{0.089} \pm 0.026$ \\ \midrule
\multirow{6}{*}{Factorio} & \multirow{3}{*}{0.00}                                      & \multicolumn{1}{p{1.1cm}|}{Success}                 & Out of mem.        & $0 \%$                                               & $0 \%$                                                & $0.0\%$                                               & N/A                                                  & N/A                                                   & N/A             & $\bm{100\%}$          \\
                          &                                                            & \multicolumn{1}{p{1.1cm}|}{$\fulllength{\upsilon}$} & --                 & --                                                   & --                                                    & --                                                    & --                                                   & --                                                    & --              & $\bm{234.0} \pm 0.0$   \\
                          &                                                            & \multicolumn{1}{p{1.1cm}|}{Time}                    & --                 & $1345 \pm 361.5$                                     & $12470 \pm 817.4$                                     & $48570\pm86.7$                                        & --                                                   & --                                                    & --              & $\bm{0.348}\pm0.156$   \\ \cmidrule(l){2-11} 
                          & \multirow{3}{*}{0.05}                                      & \multicolumn{1}{p{1.1cm}|}{Success}                 & Out of mem.        & $10 \%$                                              & $0 \%$                                                & $0.0\%$                                               & N/A                                                  & N/A                                                   & N/A             & $\bm{100\%}$          \\
                          &                                                            & \multicolumn{1}{p{1.1cm}|}{$\fulllength{\upsilon}$} & --                 & $68.0 \pm 0.0$                                       & --                                                    & --                                                    & --                                                   & --                                                    & --              & $\bm{219.4} \pm 44.82$ \\
                          &                                                            & \multicolumn{1}{p{1.1cm}|}{Time}                    & --                 & $1345 \pm 361.5$                                     & $12470 \pm 817.4$                                     & $34900\pm427.6$                                       & --                                                   & --                                                    & --              & $\bm{0.315} \pm 0.155$ \\ \bottomrule
\end{tabular}